\PassOptionsToPackage{table}{xcolor}
\documentclass[10pt, a4paper]{article}

\usepackage{lrec-coling2024} 
\usepackage{amsmath}
\usepackage{amsfonts}
\usepackage{booktabs}
\usepackage{makecell}
\usepackage{multirow}
\usepackage[table]{xcolor}
\usepackage{subfigure}
\usepackage{float}
\usepackage{soul}

\title{Improving Faithfulness of Large Language Models in Summarization via Sliding Generation and Self-Consistency}

\name{Taiji Li, Zhi Li, Yin Zhang\thanks{\textsuperscript{*}Corresponding Author: Yin Zhang.}} 

\address{College of Computer Science and Technology, Zhejiang University, China \\
         \{litaiji, zhili, zhangyin98\}@zju.edu.cn\\}

\abstract{
Despite large language models (LLMs) have demonstrated impressive performance in various tasks, they are still suffering from the factual inconsistency problem called hallucinations. For instance, LLMs occasionally generate content that diverges from source article, 
and prefer to extract information that appears at the beginning and end of the context, especially in long document summarization. 
Inspired by these findings, we propose to improve the faithfulness of LLMs in summarization by impelling them to process the entire article more fairly and faithfully. We present a novel summary generation strategy, namely \textbf{SliSum}, which exploits the ideas of 
sliding windows and self-consistency. Specifically, SliSum divides the source article into overlapping windows, and utilizes LLM to generate local summaries for the content in the windows. Finally, SliSum aggregates all local summaries using clustering and majority voting algorithm 
to produce more faithful summary of entire article. Extensive experiments demonstrate that SliSum significantly improves the faithfulness of diverse LLMs including LLaMA-2, Claude-2 and GPT-3.5 in both short and long text summarization, while maintaining their fluency and informativeness and without additional fine-tuning and resources. We further conduct qualitative and quantitative studies to investigate why SliSum works and impacts of hyperparameters in SliSum on performance.
\\ \newline \Keywords{Large Language Model, Summarization, Faithfulness} }

\begin{document}

\maketitleabstract

\section{Introduction}

Abstractive summarization aims to generate summaries that are fluent, informative, and faithful to the source articles. Benefiting from the popularity and development of large language models (LLMs), abstractive summarization has achieved remarkable progress in fluency and coherence \citep{zhang2023benchmarking, goyal2023news, zhang-etal-2023-extractive-summarization}. However, LLMs have propensity to generate content that contradicts or is not present in the source article~\citep{tam-etal-2023-evaluating, maynez-etal-2020-faithfulness, lin-etal-2022-truthfulqa, li-etal-2023-halueval}, which is commonly referred to as hallucination. Alleviating the hallucination of LLMs is a critical challenge for their reliability in real-world applications.



Many works~\citep{zhang2023sirens, zheng2023does} explore the mechanisms why LLMs exhibit hallucinations. 
\citet{liu2023lost} observe that the performance of LLMs significantly decreases as the length of input contexts increases, resulting in the hallucination phenomenon of LLMs being particularly serious in long document summarization. Furthermore, LLMs are sensitive to the order of context and are more likely to select information presented first or last, even for short contexts \citep{xie2024adaptive, wang2023large}. That is, the summaries generated by LLMs contain more content that occurs at the beginning and end of the source article, which has a detrimental effect on the summary quality of the entire article.

\begin{table}[t]
    \centering
    \setlength\tabcolsep{3pt}
    \scriptsize
    \resizebox{\linewidth}{!}{
    \begin{tabular}{|p{\linewidth}|}
    \hline
        \textbf{Article}: Albert Einstein 
        was a German-born theoretical physicist, widely held to be one of the greatest and most influential scientists of all time. Best known for developing the theory of relativity, he also made important contributions to quantum mechanics. ...

        He received the 1921 Nobel Prize in Physics "for his services to theoretical physics, and especially for his discovery of the law of the photoelectric effect", a pivotal step in the development of quantum theory. His work is also known for its influence on the philosophy of science. ...
        
        His mass–energy equivalence formula $E = mc^2$, which arises from relativity theory, has been called "the world's most famous equation". His intellectual achievements and originality have made the word Einstein broadly synonymous with genius. ...
        
        For much of the last phase of his academic life, Einstein fought a long rearguard action against quantum theory's introduction of fundamental randomness into science's picture of the world, objecting that "God does not play dice". ...\\
    \hline
        \textbf{Local Summaries}\\
    \hline
        \textbf{Paragraph 1 and 2}: Einstein is one of the greatest and most influential theoretical physicist and \textcolor{red}{\ul{won the 1921 Nobel Prize in Physics for philosophy of science}}. \\
    \hline
        \textbf{Paragraph 2 and 3}: \textcolor{red}{\ul{Einstein received the 1921 Nobel Prize in Physics for his discovery of the law of the photoelectric effect.}} \textcolor{violet}{\ul{Einstein is synonymous with genius.}}\\
    \hline
        \textbf{Paragraph 3 and 4}: \textcolor{violet}{\ul{The mass-energy equivalence formula is the most famous equation.}} Einstein objected to the randomness of quantum theory in his academic life.\\
    \hline
    \end{tabular}}
    \caption{The self-contradiction problem between local summaries. The \textcolor{red}{\underline{red sentences}} are contradictory statements about the reason why Einstein won the Nobel Prize in Physics. The \textcolor{violet}{\underline{purple sentences}} are different summaries of the third paragraph.}
    \label{contradiction}
\end{table}

Recent works attempt to leverage post-processing models and the CoT (Chain of Thought) technique to improve factual consistency of summarization. Some works \citep{liu-etal-2023-improving, xiao-etal-2024-personalized} employ LLMs as critic and editor models, the critic model generates editing instructions for the initial summary, and the editor model corrects factual errors by following the instructions. \citet{wang-etal-2023-element} propose a CoT-based method called SumCoT to elicit LLMs to generate summaries step by step. However, they do not mitigate position bias and performance degradation in long context scenarios. 

\begin{figure*}[t]
    \begin{center}
        \includegraphics[width=\linewidth]{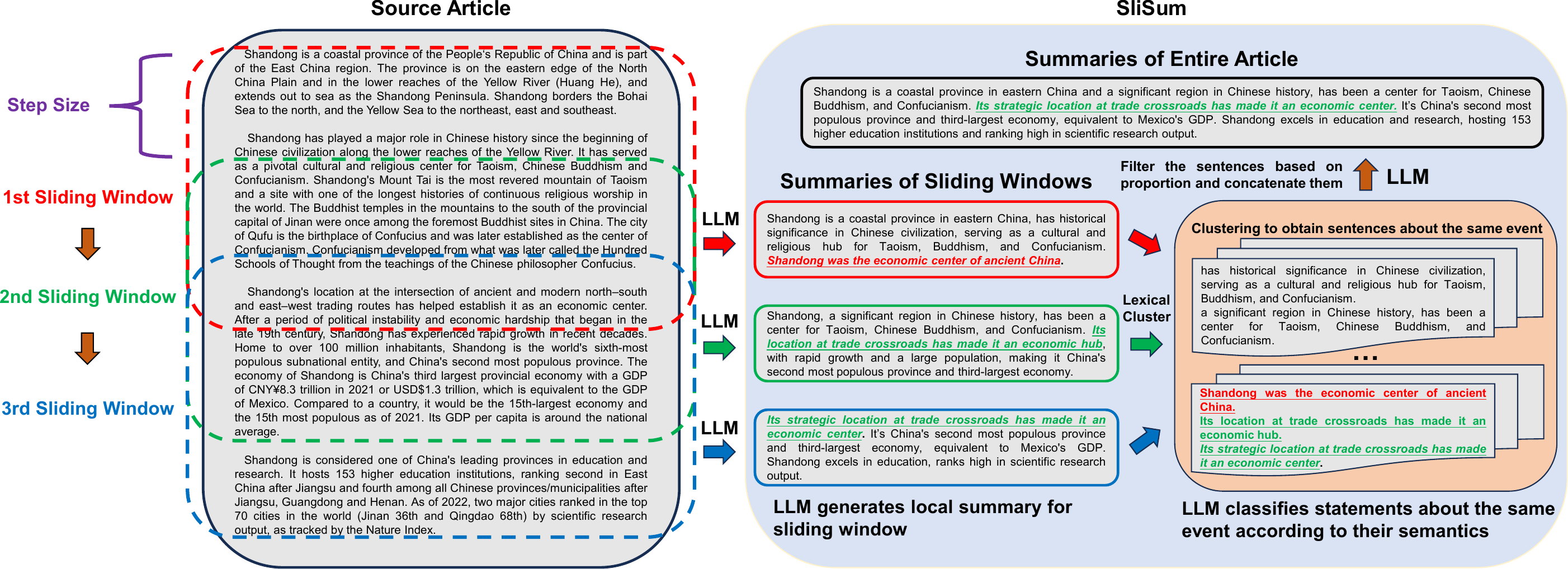} 
        \caption{The pipeline and example of our proposed \textbf{SliSum} approach. 
        In order to solve self-contradiction problem, SliSum take majority vote over sentences of each cluster base on their semantics and select the category with the most votes. For instance, the \textbf{\textcolor{green}{\underline{green sentences}}} have the similar semantics and appear twice, while the \textbf{\textcolor{red}{\underline{red sentence}}} with different semantics appear only once. Hence, the \textbf{\textcolor{green}{\underline{second green sentence}}} is selected to be output to the final summary. In the implementation, SliSum processes the source article at the sentence level. For the simplicity of the illustration, the windows in the figure are represented by text lines.}
        \label{architecture}
    \end{center}
\end{figure*}

In this paper, we propose \textbf{SliSum}, a novel summary generation strategy that improves the faithfulness of LLMs in both short and long text summarization by sliding generation and self-consistency. As shown in Figure \ref{architecture}, our approach consists of the following three steps: (1) \textbf{Sliding Generation}: SliSum uses LLM to generate local summaries for overlapping windows, and local summaries contain multiple statements about the same event that may have contradictions (see Table \ref{contradiction}).  
(2) \textbf{Filtration}: According to the principle of self-consistency, the statements generated more times by LLMs are more faithful and important to the source article~\citep{wang2023selfconsistency, manakul-etal-2023-selfcheckgpt}. 
Therefore, we obtain statements about the same event by lexical clustering over sentences of all local summaries and then filter out small sentence clusters and outliers so that the global summary contains only relatively more important information of the source article. (3) \textbf{Aggregation}: SliSum employs LLM to detect whether there are contradictions in statements about the same event and classify them into different categories based on their semantics. Finally, SliSum uses a majority voting algorithm to select the statements with the most proportion, and concatenates them to form a more faithful summary of the entire article. SliSum brings three major benefits: (1) Sliding window provides LLMs with more diverse and adequate information by splitting articles in an overlapping manner; (2) The filtration and aggregation based on self-consistency ingeniously mitigate the self-contradiction problem and further leverage the potential of LLMs to improve the faithfulness of them; (3) The combination of sliding windows and self-consistency impels LLMs to process the entire article more fairly and faithfully. Therefore, SliSum is capable of improving the faithfulness of LLMs in summarization without external resources and additional fine-tuning.

We evaluate the effectiveness of SliSum applied to three advanced LLMs, LLaMA-2-13B \citep{touvron2023llama}, Claude-2 \citep{claude} and GPT-3.5 \citep{chatgpt}, on four popular summarization datasets. Extensive experiments have shown that SliSum significantly improves the faithfulness of three LLMs on the short news datasets CNN/DM and XSum, and the long scientific papers datasets arXiv and PubMed, respectively, while without sacrificing their fluency and informativeness. Besides, we conduct ablation studies to further investigate why SliSum works. We also perform quantitative investigations on hyperparameters in SliSum. Our contributions are summarized as follows.

\begin{itemize}
    \item We propose a novel summary generation architecture, \textbf{SliSum}, that improves faithfulness of LLMs by sliding windows and self-consistency without additional resources and fine-tuning. To the best of our knowledge, we are the first to apply overlapping context windows into LLMs for abstractive summarization.
    \item We demonstrate that SliSum uniformly improves factual consistency of summaries generated by diverse LLMs while maintaining their fluency and informativeness, more importantly, SliSum is applicable to text of various lengths and styles.
    \item We conduct extensive qualitative and quantitative experiments to validate the effectiveness of sliding generation and aggregation based on self-consistency and impacts of hyperparameters in SliSum on performance.
\end{itemize}

\section{Related Works}

\paragraph{Factual Consistency of Summarization} The factual consistency in abstractive summarization has received increasing attention recently. Existing work has proposed various methods to improve the factual consistency, such as contrastive learning \citep{wan-bansal-2022-factpegasus, xie-etal-2023-alleviating}, adversarial learning \citep{wang-etal-2022-improving, wu-etal-2022-frsum}, textual entailment \citep{zhang-etal-2022-improving-faithfulness, roit-etal-2023-factually} and post-editing \citep{fabbri-etal-2022-improving, balachandran-etal-2022-correcting}. However, these methods can not be directly applied in long document summarization due to the difficulty of modeling long texts accurately. Although recent studies have addressed this problem by leveraging Graph Neural Networks \citep{zhang-etal-2022-hegel, doan-etal-2022-multi, phan-etal-2022-hetergraphlongsum, xie-etal-2022-gretel}, reinforcement learning \citep{gu-etal-2022-memsum} and structure information \citep{cao-wang-2022-hibrids, cho-etal-2022-toward, pu-etal-2023-incorporating}, they do not improve the faithfulness of long document summarization systems. There are currently few works that focus on improving factual consistency of long document summarization. In contrast, we propose a unified architecture that consistently improves factual consistency of both short and long text summarization.


\paragraph{Mitigation of LLM Hallucination} Recent studies have made several attempts to mitigate the hallucination of LLMs, including retrieval-augmented generation \citep{peng2023check, ram2023incontext, kang2023knowledge, xu2024retrieval}, post-processing models \citep{chen2023purr, gou2024critic, huang-etal-2023-zero}, prompt engineering \citep{xue2023rcot, shi2023leveraging, luo2023zeroresource, dhuliawala2023chainofverification} and self-supervised learning \citep{gekhman-etal-2023-trueteacher, manakul-etal-2023-selfcheckgpt, du2023improving}. However, most of these works require training additional models or external knowledge, or lack sufficient evaluation on long contexts tasks. For example, LLM-Augmenter \citep{peng2023check} acquires evidence by retrieving external knowledge for the LLM to generate candidate responses grounded in evidence. 
\citet{du2023improving} propose a post-processing method that improves factuality of language models using multiagent debate. SELF-FAMILIARITY \citep{luo2023zeroresource} has revealed excellent performance in short context tasks, but its advancement has not been demonstrated on long-context datasets. Different from previous works, SliSum improves faithfulness of LLMs in long context scenarios without additional fine-tuning and external resources.

\paragraph{Long Context for LLMs} With the application fields of LLMs continue to expand, boosting the performance of LLMs in long context scenarios has aroused a surge of interest. \citet{han2023lminfinite} propose a decoding method, LM-Infinite, to maintain fluency and generation quality of LLMs on long sequences, which can only be applied to LLMs using relative positional encodings, but our approach is applicable to diverse LLMs, even black-box models, because it does not modify the original structure and implementation of the models. \citet{jiang2023longllmlingua} and \citet{li-etal-2023-compressing} compress long context by removing redundancy to enhance LLMs' perception of the key information, but there is no guarantee that compressed contexts contain faithful information to original contexts. 
\citet{ratner-etal-2023-parallel} propose a parallel context windows method that LLMs process contexts located in these windows separately to mitigate the negative impact of long contexts on performance. However, parallel windows separate the semantic relationships within the article, which inevitably affects LLMs’ understanding and extraction of global information. 
We are the first to introduce overlapping sliding windows into LLMs for abstractive summarization tasks. Unlike traditional parallel windows, sliding windows overlap each other so that each segment of the source article can be located at the beginning of a certain window, allowing them to overcome their preference for position and process the entire context more fairly and faithfully. 
In previous works, the sliding window method is only applied to extractive  \citep{cui-hu-2021-sliding} and query-focused \citep{vig-etal-2022-exploring} summarization, because they do not solve the self-contradiction problem \citep{mündler2024selfcontradictory, liu-etal-2022-token} between local summaries. SliSum addresses this problem by combining sliding windows and self-consistency and achieves better performance than existing methods.

\section{Approach}

\subsection{Sliding Generation}\label{sg}

\paragraph{Sliding Window} SliSum first divide the source article $A$ into a list of sentences $[C_1, C_2, \cdots, C_n]$. The sliding window with predefined \textbf{\textit{window size}} initially consists of several consecutive sentences $[C_1, \cdots, C_t]$ at the beginning of the article. The number of words in the window only needs to be approximately equal to the window size. Figure \ref{architecture} shows that LLM generates a local summary for the content in the sliding window. SliSum restricts the attention of LLM within a window, alleviating the distraction issue caused by long contexts \citep{han2023lminfinite, NEURIPS2023_8511d06d}. 
Then the sliding window moves downward by a distance of \textbf{\textit{step size}} and LLM generates a local summary for current window again. Similarly, the distance moved does not need to be strictly equal to step size. That is, the starting index of the window advances a few sentences. Importantly, SliSum processes the source article at the sentence level. Finally, LLM iteratively generates summaries for the content in the current window until the sliding window traverses the entire article. Given an article $A = [C_1, C_2, \cdots, C_n]$, the process of sliding generation can be described as follows:

\begin{align*}
    \mathrm{S_1} & = SUM (\mathrm{W_1}) = SUM ([C_{p_1}, \cdots, C_{q_1}])\\
    \mathrm{S_2} & = SUM (\mathrm{W_2}) = SUM ([C_{p_2}, \cdots, C_{q_2}])\\
    & \vdots \\
    \mathrm{S_m} & = SUM (\mathrm{W_m}) = SUM ([C_{p_m}, \cdots, C_{q_m}])\\
\end{align*}

where $SUM(W_i)$ represents LLM summarizes the content in the window $W_i$, $W_i$ consists of from $p_i$-th to $q_i$-th sentences. $p_1  = 1$, $q_m = n$, and

$$
    p_i \leq p_{i + 1} \leq q_i, \quad i = 1, \cdots, m - 1
$$

Each text segment except the beginning and end can locate in different positions in the context window. SliSum balances the position distribution in the context window of text segments, eliminating the position preference of LLMs for source article.

\paragraph{Process Repeatedly} Eventually SliSum will obtain a concatenation of summaries $[S_1, S_2, \cdots, S_m]$ by sliding generation. In the sliding window method, step size must be less than or equal to window size, otherwise there will be gaps between windows. Figure \ref{architecture} shows that LLM processes the same part of the source article multiple times due to overlaps between windows. Generally, given window size $L_w$, step size $L_s$ ($L_s \leq L_w$) and an article whose length $L_a$ is far larger than the window size, i.e, $L_a \gg L_w$, the minimum number of times the middle part of the article is summarized by LLM:

\begin{equation}\label{equation1}
    K = \left\lfloor \frac{L_w}{L_s} \right\rfloor
\end{equation}

It can be easily seen that if $L_w$ is an integer multiple of $L_s$ in sliding window method, LLM will read the middle part of the source article the same number of times, so we set $L_w = K L_s$ in SliSum. In particular, for several windows at the beginning and end, we repeatedly generate local summaries for them so that all parts of the source article are summarized $K$ times. 

\subsection{Events Filtering}

\paragraph{Lexical Clustering} The event is usually composed of entity, behavior, reason, result and other elements, so statements about different events have their own characteristic words that can be used to distinguish them. We observe that the statements generated by LLMs for the same event are usually lexically similar. Consequently, we can obtain a set of sentences about the same event by lexical similarity clustering. As shown in Figure \ref{architecture}, SliSum divides all summaries into sentences, then uses DBSCAN \citep{dbscan} algorithm to cluster them based on lexical similarity. Specifically, we use ROUGE-1 \citep{lin-2004-rouge} F1 score to define the distance between two sentences $C_1$ and $C_2$:

\begin{equation}
    dist(C_1, C_2) = 1 - \mathcal{R}(C_1, C_2)
\end{equation}

where $\mathcal{R}(\cdot)$ is ROUGE-1 F1 score, and is computed as follow:

$$
    \mathcal{R}(C_1, C_2) = \frac{2 * \textrm{number of overlap words}}{\textrm{length of } C_1 + \textrm{length of } C_2} 
$$

Obviously, $\mathcal{R}(C_1, C_2) \in [0, 1]$. There are two important parameters in the DBSCAN algorithm: distance threshold $\boldsymbol{\varepsilon}$ and minimum number of points (\textbf{MinPts}) in a cluster. Formally, let $\hat{C}$ represents a sentence, if exists a set of sentences (including $\hat{C}$ itself) $\mathbb{C} = \{C_1, C_2, \cdots, C_N\}$ and the number of sentences $|\mathbb{C}| \geq \mathrm{MinPts}$, every sentence $C \in \mathbb{C}$ satisfies

$$
    dist(\hat{C}, C) \leq \varepsilon
$$

$\mathbb{C}$ will be considered a cluster. A lower MinPts induces the algorithm to construct more clusters, while a higher MinPts will ensure more robust and consistent clusters. 

\paragraph{Filtering Noise} Intuitively, if an event appears in the local summary of a window, it means that the event is important to the content in the window. We notice that some sentences are only important in minority windows and should not appear in the overall summary. Furthermore, LLMs occasionally generate few hallucination statements that deviate from the source article. These statements tamper with information in the source article or add information that cannot be inferred from the source article, thus they generally have low lexical similarity to factually consistent statements. These undesirable statements are treated as outliers and noise in DBSCAN algorithm. Due to MinPts determines the minimum number of points in a cluster, so we can filter out unimportant and hallucination statements by using an appropriate MinPts. 

As mentioned in Section \ref{sg}, SliSum generate $K$ summaries for the content of the source article by sliding generation, that is, sentences about the same event occur at most $K$ times. Under normal circumstances, a cluster contains at most $K$ sentences. In order to generate more concise and relative summaries, we select sentences that exist in as least MinPts local summaries, i.e, we only retains clusters whose size exceeds MinPts. Therefore, we can adjust MinPts in the range of $[1, K]$ to filter noise and hallucination as well as improve self-consistency.

\subsection{Contradictions Detection and Sentences Aggregation}

\paragraph{Sentences Selection} As shown in Table \ref{contradiction} and Figure \ref{architecture}, there may be contradictory statements in sentences about the same event. Due to LLMs have ability to detect self-contradictory statements without relying on additional knowledge \citep{mündler2024selfcontradictory}, SliSum first manipulates the LLM used in the sliding generation to divide sentences into different categories according to their semantics. Generally, the more times a certain response is output by LLMs, the higher the probability that the response is desirable \citep{wang2023selfconsistency, manakul-etal-2023-selfcheckgpt}, since self-consistency reflects factual consistency. Consequently, SliSum can utilize majority voting algorithm to select sentences that are faithful to the source article. 

After distinguishing sentences that are high lexically similar but semantically different, sentences of the same category state the same fact. SliSum selects the sentence with the largest proportion to output to the final summary. When there are clusters of sentences with the same proportion or when it is necessary to choose sentence from a certain  cluster, the relatively last generated sentence is selected by SliSum. Because the last generated sentence is located at the beginning of the corresponding window, and LLMs have the highest factual consistency for the information occurs at the beginning of the input context \citep{ravaut2024context, chhabra2024revisiting}. For example, an event has five related statements $[D_1, \cdots, D_5]$ generated sequentially, and LLM divides them into $[D_2]$, $[D_1, D_4]$, $[D_3, D_5]$, SliSum selects $D_5$ as the most faithful statement to form the summary of the entire article.

\paragraph{Sentences Integration} SliSum finally integrates all selected sentences to output the final summary of the entire article. SliSum uses LLM to generate connectives and concatenate these sentences in order, so that the order of the final summary is consistent with the source article. We determine the order of the sentences according to the order of the events they describe. In the integration stage, SliSum do not need to change any information of the selected sentences.

\begin{table}[htbp]
\centering
\small
    \begin{tabular}{p{0.93\linewidth}}
    \toprule
        \rowcolor{gray!30} \textbf{Summary Generation}: Generate local summary for content in the window.\\
    \midrule
        \underline{\textit{Instruction}}: Summarize the above article.\\ 
    \bottomrule
        \rowcolor{gray!30} \textbf{Sentence Classification}: Divide the statements about the same event.\\ 
    \midrule
        \underline{\textit{Instruction}}: Classify the above statements into different categories. Statements of the same category describe the same facts, and statements of different categories have different semantics.\\
    \bottomrule
       \rowcolor{gray!30} \textbf{Sentences Integration}:  Concatenate selected sentences in
order.\\ 
    \midrule
        \underline{\textit{Instruction}}: Generate connectives to concatenate sentences to form a fluent text. DO NOT change the original semantics.\\
    \bottomrule
    \end{tabular}
    \caption{Prompts used in SliSum.}
    \label{prompts}
\end{table}
 
\section{Experimental Setup}

\begin{table*}[t]
\centering
\resizebox{\linewidth}{!}{
    \begin{tabular}{l|cccc|cc|cccc|cc}
    \toprule
         & \large{\textbf{R-1}} & \large{\textbf{R-2}} & \large{\textbf{R-L}} & \large{\textbf{BS}} & \large{\textbf{FC}} & \large{\textbf{SC}} & \large{\textbf{R-1}} & \large{\textbf{R-2}} & \large{\textbf{R-L}} & \large{\textbf{BS}} & \large{\textbf{FC}} & \large{\textbf{SC}} \\ 
    \midrule
        \large{\textbf{Model}} & \multicolumn{6}{c|}{\large{\textbf{CNN/DM}}} & \multicolumn{6}{c}{\large{\textbf{XSum}}}\\
    \midrule
        CLIFF & 43.87 & 20.63 & 40.75 & 88.67 & 51.68 & 47.19 & 44.83 & 21.52 & 36.59 & 91.54 & 22.68 & 23.71 \\
        FES & 46.49 & 22.43 & 43.19 & 89.21 & 54.91 & 52.08 & 47.46 & 24.61 & 39.24 & \textbf{91.78} & 24.29 & 24.54\\
        \hline
        LLaMA-2 & 43.79 & 20.46 & 40.69 & 86.65 & 49.53 & 46.89 & 44.58 & 21.39 & 36.47 & 91.28 & 22.61 & 25.34\\ \rowcolor{orange!30}
        \textbf{LLaMA-2 + SliSum} & 45.37 & 22.54 & 42.19 & 88.51 & 54.81 & 51.74 & 47.94 & 24.93 & 39.26 & 91.45 & 27.83 & 25.76\\
        Claude-2 & 44.83 & 21.62 & 41.54 & 88.18 & 52.97 & 50.26 & 45.85 & 22.68 & 37.24 & 91.17 & 24.35 & 23.56\\ \rowcolor{orange!30}
        \textbf{Claude-2 + SliSum} & \textbf{47.75} & \textbf{23.16} & \textbf{44.26} & \textbf{90.17} & \textbf{60.34} & 58.19 & 48.31 & \textbf{25.48} & 40.07 & 91.46 & 27.19 & 26.84\\
        GPT-3.5 & 44.35 & 21.02 & 41.28 & 88.36 & 51.62 & 49.25 & 47.69 & 24.82 & 39.42 & 91.29 & 26.74 & 23.79\\
        GPT-3.5 + SumCoT & 45.25 & 21.72 & 41.68 & 88.61 & 52.54 & 50.87 & 47.83 & 25.17 & 39.52 & 91.02 & 27.91 & 25.48\\ \rowcolor{orange!30}
        \textbf{GPT-3.5 + SliSum} & 46.48 & 22.82 & 43.26 & 89.77 & 58.31 & \textbf{58.26} & \textbf{48.74} & 25.46 & \textbf{40.18} & 91.27 & \textbf{28.58} & \textbf{27.42}\\
    \midrule
        \large{\textbf{Model}} & \multicolumn{6}{c|}{\large{\textbf{arXiv}}} & \multicolumn{6}{c}{\large{\textbf{PubMed}}}\\
    \midrule
        FactorSum & 48.29 & 20.35 & 42.38 & 88.27 & 56.38 & 53.29 & 47.41 & 20.46 & 42.73 & 82.35 & 64.21 & 58.36\\
        Lodoss & 48.35 & 20.75 & 42.56 & 88.61 & 68.44 & 65.31 & 49.34 & 23.52 & 44.86 & 88.74 & 77.18 & 75.23\\
        \hline
        LLaMA-2 & 37.57 & 12.82 & 33.71 & 74.46 & 47.31 & 42.67 & 39.28 & 16.67 & 36.47 & 76.39 & 58.39 & 52.73\\ \rowcolor{orange!30}
        \textbf{LLaMA-2 + SliSum} & 47.23 & 19.94 & 41.29 & 85.61 & 66.39 & 64.12 & 47.84 & 20.96 & 42.85 & 84.69 & 72.54 & 69.58\\
        Claude-2 & 44.95 & 17.62 & 39.54 & 80.83 & 64.26 & 61.57 & 45.27 & 19.39 & 41.54 & 81.52 & 71.24 & 67.42\\ \rowcolor{orange!30}
        \textbf{Claude-2 + SliSum} & \textbf{50.94} & \textbf{21.86} & \textbf{45.91} & 88.46 & \textbf{77.43} & \textbf{75.25} & \textbf{51.49} & \textbf{24.58} & 46.37 & \textbf{89.62} & 81.34 & \textbf{79.82}\\
        GPT-3.5 & 42.69 & 17.28 & 38.68 & 82.47 & 64.59 & 62.33 & 43.98 & 18.56 & 39.36 & 79.45 & 67.25 & 65.49\\
        GPT-3.5 + SumCoT & 45.28 & 18.39 & 40.25 & 83.26 & 65.17 & 62.56 & 45.71 & 19.62 & 41.33 & 81.27 & 67.82 & 62.94\\
        GPT-3.5 + Refine & 47.53 & 20.26 & 41.37 & 86.74 & 65.91 & 64.31 & 48.69 & 22.75 & 43.16 & 86.27 & 72.39 & 69.84\\ \rowcolor{orange!30}
        \textbf{GPT-3.5 + SliSum} & 48.57 & 20.82 & 42.62 & \textbf{88.75} & 70.81 & 68.49 & 51.08 & 24.57 & \textbf{46.76} & 89.38 & \textbf{81.74} & 79.63\\
    \bottomrule
    \end{tabular}}
    \caption{Comprehensive evaluation results of SliSum on four datasets for factual consistency, relevance and fluency. \textbf{R-1/2/L} stands for ROUGE-1/2/L, \textbf{BS} is BERTScore, \textbf{FC} is FactCC, \textbf{SC} is SummaC. The best result per metric for each dataset is \textbf{bolded}.}
    \label{result}
\end{table*}

\subsection{Datasets}

We evaluate the performance of SliSum for short and long text summarization on four popular benchmark datasets: \textbf{CNN/DM} \citep{NIPS2015_afdec700} and \textbf{XSum} \citep{narayan-etal-2018-dont} are two widely-used short news summarization datasets. \textbf{PubMed} and \textbf{arXiv} \citep{cohan-etal-2018-discourse} are two scientific paper datasets for long document summarization, which are much longer than the common news articles. PubMed contains academic papers from the biotechnology domain, while arXiv contains papers from different scientific domains. 
Limited by the maximum number of requests and time cost, we randomly select 100 articles from test set of each dataset to construct our test set, respectively.


\subsection{Baselines}

To validate the effectiveness of our proposed approach, we apply SliSum to three state-of-the-art instruction-tuned LLMs, including \textbf{LLaMA-2 13B Chat}\footnote{\href{https://ai.meta.com/resources/models-and-libraries/llama/}{https://ai.meta.com/resources/models-and-libraries/llama/}}, \textbf{Claude-2}\footnote{\href{https://www.anthropic.com/index/claude-2}{https://www.anthropic.com/index/claude-2}} and \textbf{GPT-3.5-Turbo-0613}\footnote{\href{https://openai.com/blog/chatgpt}{https://openai.com/blog/chatgpt}}. 


We also compare SliSum with recent faithfulness enhancement summarization models on CNN/DM and XSum, such as \textbf{CLIFF} \citep{cao-wang-2021-cliff} and \textbf{FES} \citep{NEURIPS2022_9b6d7202}. We select GPT-3.5 with \textbf{SumCoT} \citep{wang-etal-2023-element} as LLM news summarization baseline. For arXiv and PubMed, we compare SliSum with two long document summarization baselines: \textbf{FactorSum} \citep{fonseca-etal-2022-factorizing}, a factorized energy-based abstractive model that improves the performance and applicability by separate budget decisions from selecting important content in the document, and \textbf{Lodoss} \citep{cho-etal-2022-toward}, an extractive architecture that learns robust sentence representations by performing summarization and segmentation simultaneously. We compare SliSum with \textbf{Refine} \citep{refine}, a LLM-based long document summarization method that joins the summary of the previous text segment with the next segment, and then iteratively generate summaries for them.



\subsection{Metrics}

We evaluate the factual consistency, fluency and informativeness of summaries using four different metrics: (1) \textbf{FactCC} \citep{kryscinski-etal-2020-evaluating}, a weakly-supervised, model-based approach for verifying factual consistency and identifying conflicts between source documents and generated summaries, (2) \textbf{SummaC} \citep{laban-etal-2022-summac} that enables natural language inference models to detect inconsistency, (3) \textbf{ROUGE}\cite{lin-2004-rouge}, an automatic evaluation metric for the informativeness and fluency of a summary based on lexical overlap, and (4) \textbf{BERTScore} \citep{Zhang2020BERTScore}, that computes a similarity score between candidate and reference summaries using contextual embeddings.

\subsection{Implementation}\label{hyperparameters}

We run LLaMA-2-13B with Text Generation Inference\footnote{\href{https://github.com/huggingface/text-generation-inference}{https://github.com/huggingface/text-generation-inference}} on 8 $\times$ 24GB NVIDIA GeForce RTX 3090 GPUs. For CNN/DM and XSum, we cluster and filter sentences using the DBSCAN algorithm with $\varepsilon$ = 0.25 and MinPts = 2. The window size $L_w$ is 150 words, and the step size $L_s$ is 50 words. For arXiv and PubMed, we set distance threshold $\varepsilon$ to 0.25, and MinPts to 3. The window size $L_w$ is 750 words, and the step size $L_s$ is 150 words. For SumCoT, we use the same prompts in paper. We set chunk size to 750 words in Refine. The prompts used for summary generation, sentences classification and sentences integration are listed in Table \ref{prompts}. Regarding other baselines used in the experiments, we use standard checkpoints provided by the authors and adopt the same configuration as in the corresponding papers, respectively.



\section{Results}

\subsection{Main Results}


\paragraph{Overall Performance.}The experimental results of SliSum and baselines on four summarization datasets are reported in Table \ref{result}. 
In terms of short text summarization, Claude-2 with SliSum achieves a relative gain of 13.9\% on FactCC and 15.8\% on SummaC for CNN/DM respectively. Notably, Claude-2 with SliSum also performs the best on both ROUGE and BERTScore compared with other baselines, indicating that SliSum simultaneously improves informativeness and fluency of summaries. 

For long document summarization, Claude-2 with SliSum achieves a notable relative gain of 20.5\% on FactCC and 22.2\% on SummaC for arXiv respectively. Although FactorSum and Lodoss achieve close performance to SliSum on reference-based similarity metrics, they are significantly lower than SliSum on FactCC and SummaC. Regarding PubMed dataset, SliSum enables Claude-2 and GPT-3.5 to achieve almost the same excellent performance, and outperform other baselines on all metrics. 

Overall, LLM combined with SliSum uniformly outperforms base model itself with a wide margin on all metrics, which demonstrates the generality and effectiveness of SliSum. Specially, among three LLMs, SliSum reveals the most outstanding performance on Claude-2 and the greatest gain on LLaMA-2. Besides, SliSum applied to GPT-3.5 outperforms SumCoT and Refine on all four datasets. This shows that the superiority of our approach to other LLM-based baselines. We observe that SliSum achieves higher improvement on arXiv and PubMed than CNN/DM and XSum, and leads to greater performance gain on faithfulness metrics than on informativeness metrics. Therefore, our approach can substantially improves faithfulness of various LLMs in short and long text summarization, while maintaining their fluency and informativeness.

\begin{figure}[thbp]
    \begin{center}
        \includegraphics[width=\linewidth]{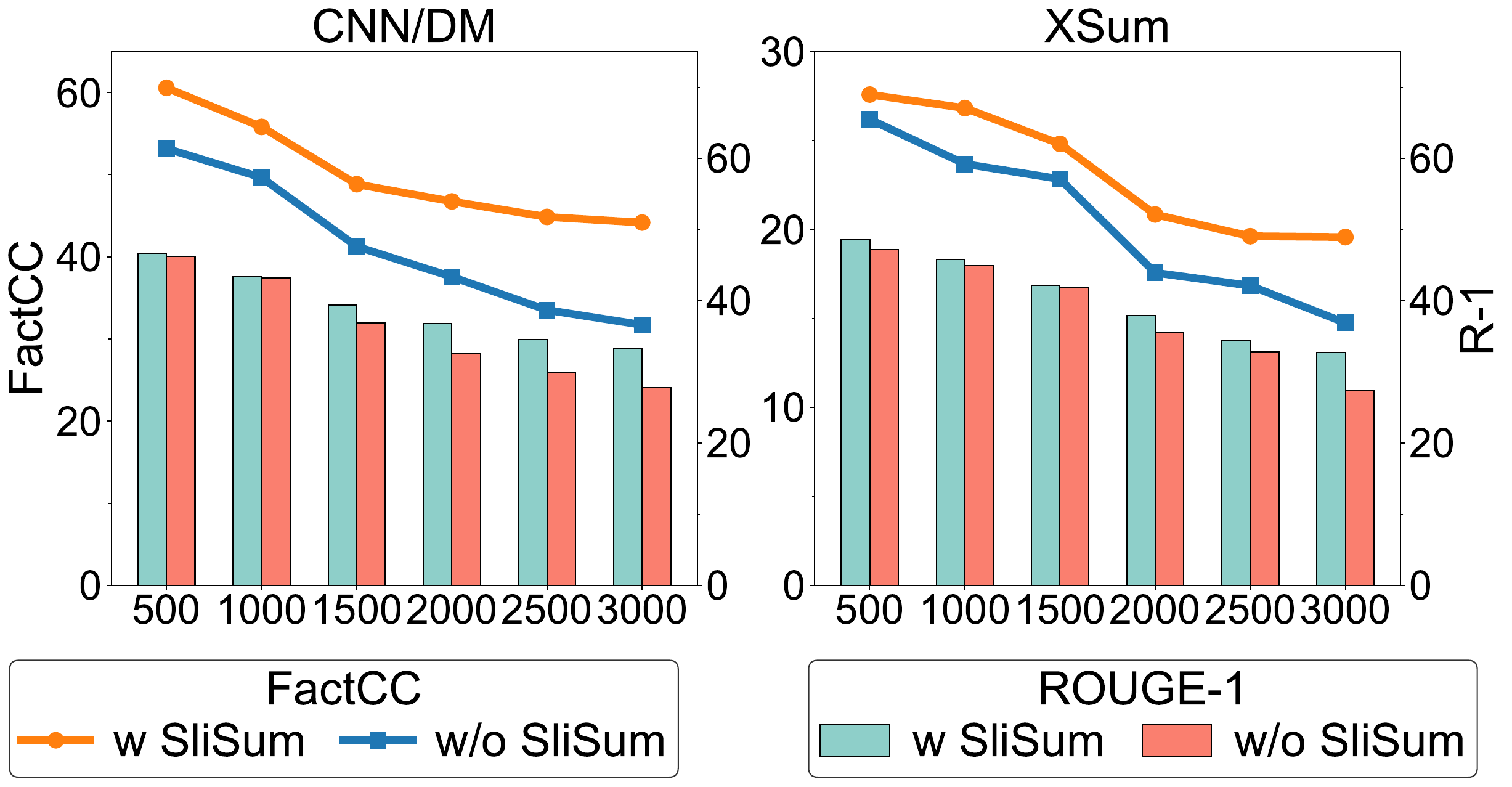}        \caption{The performance of GPT-3.5 evaluated on samples of different length.}
        \label{length}
    \end{center}
\end{figure}

\paragraph{SliSum enhances the ability to process long contexts.} We observe that SliSum has different performance gains on different datasets, which may be due to the style or length of the text. To eliminate the distractor of the text style and further demonstrate the effectiveness of our approach on news texts. We evaluate the factual consistency of GPT-3.5 on test samples of length ranges from 500 to 3000. We select 25 samples from CNN/DM and XSum with a length that is less than 5\% different from the specified length to construct the test sets, respectively. 

As shown in Figure \ref{length}, compared to the original model, SliSum significantly improves the faithfulness of GPT-3.5 in news texts of various length, and the performance gain is positively correlated with the length. This is consistent with the results in Table \ref{result} and indicates that SliSum is capable of leading to larger benefits from increasingly long context. Besides, SliSum prevents the FactCC and R-1 score of the summaries from continually decreasing as source articles grow longer. For example, when the length of articles increases from 1000 to 3000, the FactCC score of summaries generated for CNN/DM by SliSum remains stable, while the scores of the baseline still decreases. We demonstrate that our approach enhances the ability to handle long texts in a variety of styles.

\subsection{Ablation Study}

SliSum contains two essential modifications: sliding generation, filteration and aggregation based on self-consistency. In order to investigate the effect of modification designed in the SliSum, we conduct the following ablation studies by comparing SliSum with corresponding variation on GPT-3.5. 

\begin{table}[htbp]
\centering
    \resizebox{\linewidth}{!}{
    \begin{tabular}{cccccc}
        \toprule
            \textbf{Dataset} & \textbf{Method} & \textbf{R-1} & \textbf{R-L} & \textbf{FC} & \textbf{SC}\\
        \midrule
        \multirow{3}{*}{PubMed} & Single & 44.76 & 40.29 & 70.35 & 67.41\\
        & Parallel & 47.38 & 42.72 & 74.36 & 71.84\\
        & \cellcolor{orange!30}\textbf{Sliding} & \cellcolor{orange!30}\textbf{51.08} & \cellcolor{orange!30}\textbf{46.76} & \cellcolor{orange!30}\textbf{81.74} & \cellcolor{orange!30}\textbf{79.63}\\
        \bottomrule
        \\
        \toprule
        & \textbf{Position} & \footnotesize{\textbf{1-1000}} & \footnotesize{\textbf{1001-2000}} & \footnotesize{\textbf{2001-3000}} & \footnotesize{\textbf{3001-}}\\
        \midrule
        \multirow{2}{*}{arXiv} & Single & 39.26\%  & 24.63\% & 13.58\% & 22.53\% \\
        & \cellcolor{orange!30}\textbf{Sliding} & \cellcolor{orange!30}29.61\% & \cellcolor{orange!30}26.48\% & \cellcolor{orange!30}18.83\% & \cellcolor{orange!30}25.08\% \\
        \midrule
        \multirow{2}{*}{PubMed} & Single & 41.83\%  & 14.74\% & 11.29\% & 32.11\% \\
        & \cellcolor{orange!30}\textbf{Sliding} & \cellcolor{orange!30}32.96\% & \cellcolor{orange!30}21.57\% & \cellcolor{orange!30}18.32\% & \cellcolor{orange!30}27.15\% \\
        \bottomrule
    \end{tabular}}
    \caption{Evaluation results of different context processing method, and the position distribution of the generated sentences in the source article. For SW method, we use the same parameters with Section \ref{hyperparameters}. For PW method, we only modify $L_s$ equal to $L_w$.}
    \label{ablation1}
\end{table}


\begin{figure}[htbp]
    \begin{center}
        \includegraphics[width=0.9\linewidth]{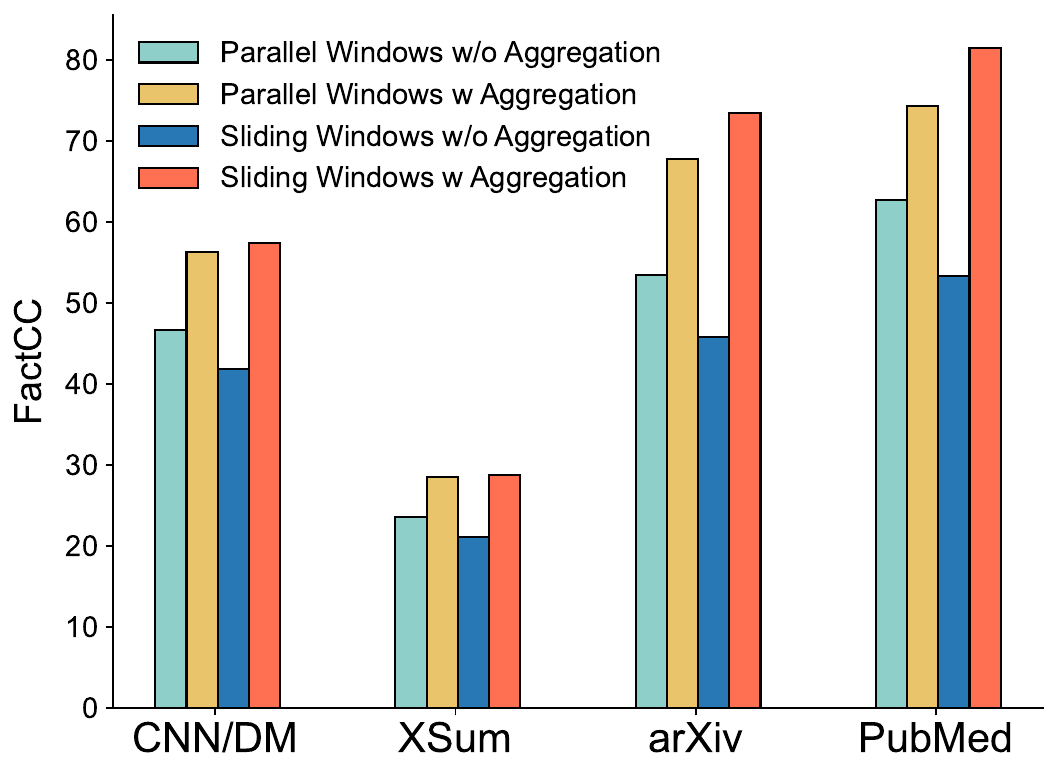} 
        \caption{The factual consistency of sliding generation with aggregation and without aggregation.}
    \label{filteration}
    \end{center}
\end{figure}

\paragraph{Sliding Windows vs. Parallel Windows.} Parallel window(\textbf{PW}), a special case of sliding window(\textbf{SW}), partitions contexts in a non-overlapping manner. To investigate the contribution of sliding generation to SliSum, we compare comprehensive performance of SW and other baselines (also generate the summary $K$ times to perform aggregation). Table \ref{ablation1} shows that single window method is worse than others on PubMed, due to the distraction issue caused by long texts. In contrast, SW outperforms PW method on all three metrics, which demonstrates that sliding generation brings performance improvement to LLMs. This is because SW can take advantage of longer context to support generating summaries, whereas PW lack direct interaction between windows, resulting in the inability to adequately extract global important information. Specifically, benefiting from overlapping windows, SW can utilize information of  $L_w + L_s(\lceil L_w / L_s \rceil - 1)$ words during generating summary for each text segment, while PW can only utilize $L_w$ words. Furthermore, Table \ref{ablation1} shows that sliding windows facilitates LLMs to more fairly and faithfully process contexts of various length without position bias. The summaries generated by SliSum contain the content of each position range of the source article relatively evenly.

\paragraph{Filteration and Aggregation improve faithfulness.} In order to understand the importance of filteration and aggregation based on self-consistency, we conduct ablation study by comparing SliSum to variation w/o aggregation. We remove two steps by directly generating global summaries for all local summaries (\textbf{Map-Reduce} \citep{mapreduce}). As shown in Figure \ref{filteration}, the FactCC scores of both SW and PW w/o aggregation are lower than their counterparts. Notably, sliding windows w/o aggregation substantially diminish the factual consistency of summaries, even worse than parallel windows w/o aggregation. This indicates that aggregation is essential for SliSum, because sliding generation while providing more diverse information also leads to self-contradiction problem, which asks us to address this problem by filteration and aggregation based on self-consistency. 

\subsection{Impacts of Hyperparameters}

We conduct quantitative experiments to investigate impacts of hyperparameters in SliSum on performance. Limited by computational resources and budget, we randomly select 25 articles from corresponding dataset to construct reduced test sets.

\begin{figure}[htbp]
    \begin{center}
        \includegraphics[width=\linewidth]{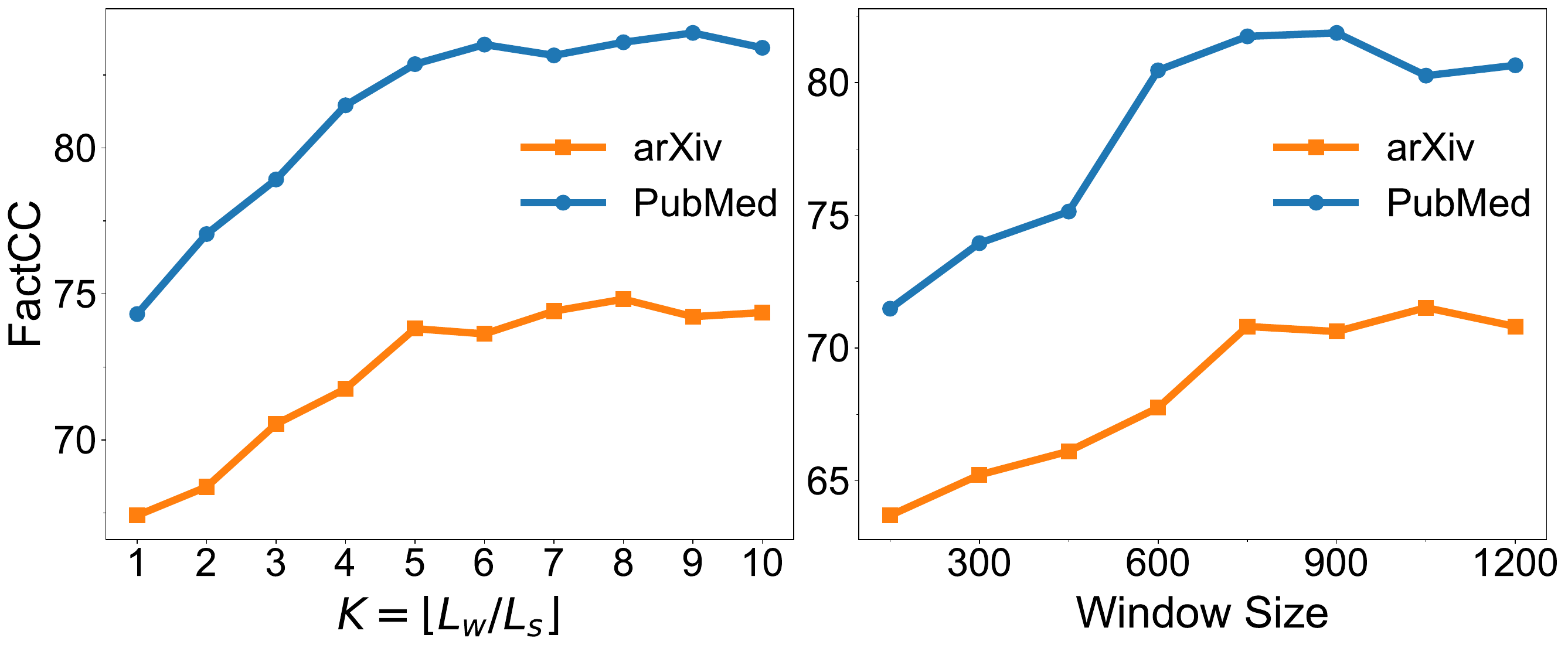} 
        \caption{Impact of ratio $L_w / L_s$ (left) and window size (right) on faithfulness of GPT-3.5. To analyze ratio, We fix $L_w = 900$ and gradually decrease $L_s$ to adjust the ratio. To analyze window size, We fix $L_w / L_s = 5$ and increase $L_w$ 150 words each time from 150 to 1200 words. }
        \label{window}
    \end{center}
\end{figure}

\paragraph{Ratio of $L_w$ to $L_s$.} Figure \ref{window} (left) shows that the FactCC score when varying in ratio $K = \left\lfloor L_w / L_s\right\rfloor$ from 1 to 10. As shown in Equation \ref{equation1}, $K$ represents the minimum number of process content of the article. A large $K$ can more adequately exploit the self-consistency of LLMs to improve performance. When the ratio grows from 1 to 5, the FactCC scores increased by 9.5\% and 11.5\% on arXiv and PubMed respectively, yet gradually converges when the ratio grows from 5 to 10, which indicates that processing articles too many times is extremely expensive and unnecessary. To save computational resources, we set the $L_w / L_s$ to 5 in our experiments.


\paragraph{Window Size.} Intuitively, shorter window means that LLMs can more accurately exploit information in context. But this has harmful effect on global understanding of entire article, resulting in the summary only contains locally important content. Nevertheless, a longer window will cause distraction issue and more uncertainty. As shown in Figure \ref{window} (right), the FactCC score of summaries initially raises as the window size grows. However, when window size increases from 750 to 1200, the performance remains stable or slightly drops. The quantitative experiments results indicate that LLMs require sufficient information to generate high-quality summaries from context, but a larger window will not further boost performance. 


\begin{table}[htbp]
\centering
    \small
    \begin{tabular}{c|cc|cc}
        \toprule
            \textbf{Dataset} & \multicolumn{2}{c|}{\textbf{arXiv}} & \multicolumn{2}{c}{\textbf{PubMed}}\\
        \midrule
            \textbf{MinPts} & \textbf{R-1} & \textbf{FC} & \textbf{R-1} & \textbf{FC}\\
        \midrule
            1 & 35.47 & 59.35 & 37.59 & 63.79\\
            2 & 41.52 & 65.86 & 43.16 & 71.46\\
            3 & \cellcolor{orange!30} \textbf{48.63} & 71.44 & \cellcolor{orange!30} \textbf{51.38} & 81.57\\
            4 & 43.18 & 75.12 & 45.62 & 83.65\\
            5 & 29.41 & \cellcolor{orange!30} \textbf{76.31} & 32.74 & \cellcolor{orange!30} \textbf{83.92}\\
        \bottomrule
    \end{tabular}
    \caption{The performance of SliSum with varying MinPts.}
    \label{minpts}
\end{table}

\paragraph{Minimum Number of Clusters (MinPts).} MinPts controls clustering and filtering of sentences about the same event. A higher MinPts forces SliSum to select sentences that are important in more windows. Table \ref{minpts} shows the FactCC and R-1 scores significantly increase with a bigger MinPts, but it will also lower R-1 score when MinPts greater than 3,  as a result of many relatively important sentences are filtered out. Therefor, we set MinPts to 3 for arXiv and PubMed. More generally, considering the balance between faithfulness and informativeness of summaries, the MinPts is set to $\frac{1}{2} K$ for different parameter configurations.

\begin{table}[htbp]
\centering
    \begin{tabular}{c|ccc}
        \toprule
            \textbf{Dataset} & \textbf{Avg} & \textbf{Max} & \textbf{Hausdorff} \\
        \midrule
            CNN/DM & 0.257 & 0.426 & 0.758\\
            XSum & 0.248 & 0.395 & 0.721\\
            arXiv & 0.253 & 0.418 & 0.764\\
            PubMed & 0.246 & 0.409 & 0.743\\            
        \bottomrule
    \end{tabular}
    \caption{The distribution of all statements in local summaries.}
    \label{distance}
\end{table}

\paragraph{Distance Threshold $\varepsilon$} The choice of distance threshold $\varepsilon$ depends on the data distribution. We count the average and the maximum distance between statements about the same event on the four datasets. Besides, we also calculate the average Hausdorff distance between sets of statements about different events. Given the set of statements $X$ and $Y$ for events A and B, the Hausdorff distance between $X$ and $Y$ is

$$d_H(X, Y) = \max\{\sup_{x \in X}{\inf_{y \in Y}{d(x, y)}}, \sup_{y \in Y}{\inf_{x \in X}{d(x, y)}}\}$$

Table \ref{distance} shows that the average distance between statements about the same event is around 0.25, which is much smaller than the distance between statements about different events. Hence, we set $\varepsilon$ to 0.25 for all datasets. The results in table \ref{distance} are consistent with our observation that statements generated by LLMs for the same event are lexically similar.

\subsection{Complexity Analysis}

\paragraph{Theoretical Analysis} SliSum reduces the computational complexity in the process of summary generation compared with the base model. For the input of length $L$, the computational complexity of most LLMs (such as LLaMA) is $O(L^2)$. However, after applying SliSum to base models, the computational complexity of summary generation decreases to $O(L)$. 
    
Generally, given window size $L_w$, step size $L_s$ ($L_s \leq L_w$) and an article whose length is $L$ is far larger than the window size, $K = L_w / L_s$ (see Equation \ref{equation1}). Our approach needs to process a total of $(L / L_s + (K - 1)^2)$ text segments of length $L_w$, hence, the complexity of SliSum is 

\begin{equation}
    O(K \times L_w \times L + (K - 1)^2 \times L_w^2) = O(L)
\end{equation}

By solving the equation, we can see that when the input length is greater than $1.36 \times K \times L_w$, the computational cost of summary generation of SliSum is less than base model.

Of course, SliSum also includes filteration and aggregation, which brings additional computational costs. However, the DBSCAN algorithm runs very fast. As for the final aggregation process, LLM usually only needs to process very few sentences. Importantly, SliSum splits the original task into smaller subtasks, so parallelization can be used to speed up inference. Overall, SliSum slightly increases the computational cost compared to the base model.

\begin{table}[htbp]
\centering
\resizebox{\linewidth}{!}{
    \begin{tabular}{c|cccc}
        \toprule
            \textbf{Dataset} & \textbf{CNN/DM} & \textbf{XSum} & \textbf{arXiv} & \textbf{PubMed}\\
        \midrule
            LLaMA-2 & 1.78 & 1.06 & 11.83 & 7.19\\
            LLaMA-2 + SliSum & 3.41 & 2.27 & 17.68 & 11.75\\
        \bottomrule
    \end{tabular}}
    \caption{The time cost (min) of LLaMA-2-13B with and without SliSum.}
    \label{time}
\end{table}

\paragraph{Quantitative Experiments} In order to verify the above theoretical studies, we recorded the inference time of the LLaMA-2-13B on four datasets (including 100 articles) to evaluate the computational cost. Table \ref{time} indicates the LLaMA-2-13B with SliSum only took twice as long as the LLaMA-2-13B without SliSum. By contrast, as shown in Table \ref{result}, SliSum substantially improves faithfulness of various LLMs. Therefore, SliSum slightly increases the computational cost, but brings relatively high performance gains.

\section{Conclusion}

In this paper, we propose \textbf{SliSum}, a novel LLM summary generation architecture, which improves faithfulness of LLMs in both short and long text summarization without additional resources and fine-tuning. SliSum leverages sliding generation as instruments for exploiting the self-consistency of LLMs, enables LLMs to more fairly and faithfully process contexts of various length, and solve contradictions between local summaries by clustering and filtering. Extensive experiments demonstrate the effectiveness of SliSum in diverse LLMs. SliSum empowers LLMs to generate faithful summaries while maintaining their fluency and informativeness. Furthermore, we conduct ablation studies to justify the modification made in the SliSum and investigate the impact of hyperparameters in SliSum. We also demonstrate through theoretical analysis and quantitative experiments that SliSum only slightly increases the computational cost.

\section*{Acknowledgements}
This work was supported by the Zhejiang Provincial Natural Science Foundation of China under Grant No. LZ23F020009, the NSFC project (No.~62072399), MoE Engineering Research Center of Digital Library, China Research Centre on Data and Knowledge for Engineering Sciences and Technology, and the Fundamental Research Funds for the Central Universities. We also express our sincere gratitude to anonymous reviewers for their invaluable feedback and constructive comments.

\section{Bibliographical References}

\bibliographystyle{lrec-coling2024-natbib}
\bibliography{SliSum}

\end{document}